\documentclass[format=acmsmall, review=false, screen=true]{acmart}

\usepackage{booktabs} 
\usepackage{graphicx}
\usepackage{amsmath}
\usepackage{courier}
\usepackage{bm}
\usepackage{bbm}
\usepackage{bbold}
\usepackage{multirow}
\usepackage{makecell}
\usepackage{mathrsfs}
\usepackage[american]{babel}
\usepackage{microtype}
\usepackage{float}
\usepackage{dblfloatfix}
\usepackage{hyperref}

\makeatletter
\def\UrlAlphabet{%
	\do\a\do\b\do\c\do\d\do\e\do\f\do\g\do\h\do\i\do\j%
	\do\k\do\l\do\m\do\n\do\o\do\p\do\q\do\r\do\s\do\t%
	\do\u\do\v\do\w\do\x\do\y\do\z\do\A\do\B\do\C\do\D%
	\do\E\do\F\do\G\do\H\do\I\do\J\do\K\do\L\do\M\do\N%
	\do\O\do\P\do\Q\do\R\do\S\do\T\do\U\do\V\do\W\do\X%
	\do\Y\do\Z}
\def\UrlDigits{\do\1\do\2\do\3\do\4\do\5\do\6\do\7\do\8\do\9\do\0}
\g@addto@macro{\UrlBreaks}{\UrlOrds}
\g@addto@macro{\UrlBreaks}{\UrlAlphabet}
\g@addto@macro{\UrlBreaks}{\UrlDigits}
\makeatother

\usepackage[ruled]{algorithm2e}

\SetAlFnt{\small}
\SetAlCapFnt{\small}
\SetAlCapNameFnt{\small}
\SetAlCapHSkip{0pt}
\IncMargin{-\parindent}

\begin{document}
	
\setcopyright{acmcopyright}
\acmJournal{TIST}
\acmYear{2019} \acmVolume{1} \acmNumber{1} \acmArticle{1} \acmMonth{1} \acmPrice{15.00}

\title[Distributed Deep Forest]{Distributed Deep Forest and its Application to Automatic Detection of Cash-out Fraud}

\author{Ya-Lin Zhang, Jun Zhou}
\affiliation{
  \institution{Ant Financial Services Group}
  \country{China}}
\email{{lyn.zyl,jun.zhoujun}@antfin.com}

\author{Wenhao Zheng, Ji Feng}
\affiliation{
	\institution{National Key Lab for Novel Software Technology, Nanjing University}
	\country{China}}
\email{{zhengwh,fengj}@lamda.nju.edu.cn}

\author{Longfei Li, Ziqi Liu}
\affiliation{
	\institution{Ant Financial Services Group}
	\country{China}}
\email{{longyao.llf,ziqiliu}@antfin.com}

\author{Ming Li}
\affiliation{
	\institution{National Key Lab for Novel Software Technology, Nanjing University}
	\country{China}}
\email{lim@lamda.nju.edu.cn}

\author{Zhiqiang Zhang, Chaochao Chen, Xiaolong Li, Yuan (Alan) Qi}
\affiliation{
	\institution{Ant Financial Services Group}
	\country{China}}
\email{{lingyao.zzq,chaochao.ccc,xl.li,yuan.qi}@antfin.com}

\author{Zhi-Hua Zhou}
\affiliation{
	\institution{National Key Lab for Novel Software Technology, Nanjing University}
	\country{China}}
\email{zhouzh@lamda.nju.edu.cn}

\begin{abstract}
Internet companies are facing the need for handling large-scale machine learning applications on a daily basis and distributed implementation of machine learning algorithms which can handle extra-large scale tasks with great performance is widely needed. Deep forest is a recently proposed deep learning framework which uses tree ensembles as its building blocks and it has achieved highly competitive results on various domains of tasks. However, it has not been tested on extremely large scale tasks. In this work, based on our parameter server system, we developed the distributed version of deep forest. To meet the need for real-world tasks, many improvements are introduced to the original deep forest model, including MART (Multiple Additive Regression Tree) as base learners for efficiency and effectiveness consideration, the cost-based method for handling prevalent class-imbalanced data, MART based feature selection for high dimension data and different evaluation metrics for automatically determining of the cascade level. We tested the deep forest model on an extra-large scale task, i.e., automatic detection of cash-out fraud, with more than 100 millions of training samples. Experimental results showed that the deep forest model has the best performance according to the evaluation metrics from different perspectives even with very little effort for parameter tuning. This model can block fraud transactions in a large amount of money each day. Even compared with the best-deployed model, the deep forest model can additionally bring into a significant decrease in economic loss each day.

\end{abstract}



\maketitle

\renewcommand{\shortauthors}{Y.-L. Zhang et al.}

\section{Introduction}

Internet companies such as Ant Financial and Alibaba are facing the pressing need of developing algorithms for large scale machine learning applications, such as recommendation system \cite{davidson2010youtube,pazzani2007content,he2014practical}, advertising \cite{lacerda2006learning,ribeiro2005impedance}, and fraud detection \cite{chandola2009anomaly,zhang2018anomaly}, etc., among which the detection of cash-out fraud for the financial services is a non-negligible task. 
Cash-out fraud is getting to be a severe threat for online credit financial firms like Ant Financial. The typical procedure of cash-out fraud can be described as the following:
The shopper scans a seller-provided QR code with Alipay and pays to the seller using Ant Credit Pay (a credit service provided by Ant Financial, users can consume with the credit and pay back later), and then the shopper gets cash from the seller, and the seller can get a reward from the shopper for helping him to conduct this abnormal transaction. In this process, the shopper gets cash by consuming his credit but not money in Ant Credit Pay, and he is not aimed at buying something but to get the cash by consuming his credit. If he doesn't pay back later, this will result in an economic loss for the company. 

Without a proper strategy to detect and prevent fraud behavior, a huge amount of money may lose each day from cash-out fraud, making it a serious threat. To handle this problem, more and more companies are trying the techniques of machine learning to automatically detect the potential fraud transaction.
Nowadays, machine learning based methods such as logistic regression (LR) ~\cite{hosmer2013applied} and multiple additive regression trees (MART) ~\cite{friedman2001greedy} are widely employed, and these methods have brought into some improvement for this task. However, a more effective machine learning algorithm is always with strong demand, since this task is closely connected with the economy, and numerous transactions are conducted by Ant Credit Pay each day. Although the amount of fraudulent transactions is only a tiny percentage of the amount of all transactions, the fraudulent transactions will still lead to serious economic loss, and a small improvement with the machine learning model means that a large amount of cash-out fraud will be detected and will bring into an obvious decrease in economic loss. Thus, the exploration and deployment of new machine learning approaches for this task is always an important issue for these companies.

Deep forest is a recently proposed approach which opens a new way of building deep models with non-differentiable components, especially tree ensembles~\cite{zhou2017deep,zhou2018deep}. This new kind of deep model is shown to be able to achieve the best performance among all non-DNN methods and give competitive results with state-of-arts DNN models in a variety of domain of tasks, including face recognition, image categorization, music classification, sentiment classification, and some low dimension data, etc. 
In addition, the number of layers in the deep forest can be automatically determined according to the metric evaluated, making the model complexity being adaptive with exact data (other than a pre-defined DNN structure). What's more, the deep forest approach has much fewer hyper-parameters to tune when comparing to DNN models. In fact, according to the paper, a default setting will produce highly competitive results across all these aforementioned different tasks, making it a good candidate for an off-the-shelf classifier. 
However, the effectiveness of this model has not been validated for extra-large scale industrial tasks. Many features of industrial tasks, such as high-dimension, class-imbalance, and extremely large scale need to be taken into consideration for the employment of this model in industrial tasks. The distributed version of the deep forest model is needed for handling industrial tasks.

In this work, based on the parameter server based system KunPeng~\cite{zhou2017kunpeng,zhou2017psmart}, we implemented the distributed version of the deep forest model with industrial standard, which is able to handle millions of high-dimensional data. 
To meet the need for industrial tasks, many improvements have been introduced for our developed deep forest model based on the original version of the deep forest model. To name a few, MART is employed as the base learner for the consideration of both efficiency and effectiveness, the cost-based strategy is applied for handling extra-imbalanced data,  feature selection with MART is adopted for high dimension data, different evaluation metrics are provided for automatically determining of the layer amount in the cascade level.

We validate the performance of the deep forest model on the crucial task of automatic detection of cash-out fraud, which is with an extremely large scale and severely class-imbalanced. The results show that the performance of deep forest is significantly better than all previously deployed methods with regard to different evaluation metrics. Deep forest is able to block fraud transactions in a large amount of money per day. What's more, the robustness of deep forest is also verified through the experiments.

Briefly speaking, the main contribution of this work can be concluded as follows: 
\begin{description}
	\item[ $\bullet$ ] We implement and deploy the first distributed version of the deep forest model based on the existing distributed system KunPeng.
	\item[ $\bullet$ ] Many improvements are brought based on the original version of the deep forest model, which includes MART as base learners for efficiency and effectiveness consideration, the cost-based method for handling prevalent class-imbalanced data, MART based feature selection for high dimension data and different evaluation metrics for automatically determining of the cascade level.
	\item[ $\bullet$ ] We validate the performance of the deep forest model on a crucial task named automatic detection of cash-out fraud, which is with extremely large scale and class-imbalanced, the results show that the performance of deep forest is significantly better than all existing methods with regard of different evaluation metrics. Besides, the robustness of deep forest is verified through the experiments too.
\end{description}

The rest of the paper is organized as follows: First, we give a detailed description of the whole system, in which the improvements of the specified deep forest model are addressed. Then, experiments on the task of automatic detection of cash-out fraud are presented and results from different perspectives are analyzed. Finally, we conclude this paper and discuss for future work.

\section{System Overview}
In this section, we give a detailed description of the whole system. Since the system is based on our distributed system KunPeng, we will first briefly introduce the system. Then, we discuss the distributed MART since it is used as the base learner for the deep forest. After that, the specific deep forest model along with the improvements are addressed, which is important for the deployment of this model in an industrial standard. The following part is about distributed implementation and job scheduling. Finally, an easy-to-use graphical user interface is introduced.

\subsection{KunPeng System}
KunPeng \cite{zhou2017kunpeng,zhou2017psmart} is a parameter server based distributed learning system with parallel optimization algorithms developed to handle the large-scale problems that arise in the industrial community, and it is among the world's largest online learning systems which can process data at petabyte and models with billions of parameters. A parameter server system \cite{dean2012large,li2014scaling,xing2015petuum} is composed of two main parts: the stateless workers, which performs the bulk of computation tasks of model training, and the stateful servers which maintain the parameters of the model. The huge model parameters are distributed on the servers and can be passed to the workers through network communication so that hundreds of billions of model parameters can be handled. Besides, the parameter server also provides a solution to the node failures in the clusters with the help of checkpoints. 

Due to these advantages, KunPeng is developed, as a production-level parameter server based distributed learning system. Briefly speaking, KunPeng is built with many optimizations: (1) A robust failover mechanism which guarantees the high success rate of large-scale jobs; (2) An efficient communication implementation for sparse data and general communication interfaces; (3) A user-friendly C++ and Python SDKs \cite{zhou2017kunpeng}.

Many popular algorithms are implemented, such as the Follow- the-Regularized-Leader Proximal (FTRL-Proximal)~\cite{mcmahan2013ad}, Multiple Additive Regression Trees (MART) algorithm~\cite{friedman2001greedy} and its extension LambdaMART~\cite{burges2010ranknet}, the Sparse Logistic Regression algorithm~\cite{liu2009large}, Factorization Machines~\cite{li2016difacto}, Latent Dirichlet Allocation (LDA) algorithm~\cite{blei2003latent} and deep learning~\cite{goodfellow2016deep} framework based on CPU cluster, and so on. Based on this system, many other algorithms can be further developed to handle extremely large-scale tasks.

\begin{figure}[htbp]
	\centering
	\includegraphics[width=10cm,trim={3cm 0cm 4cm 0cm},clip]{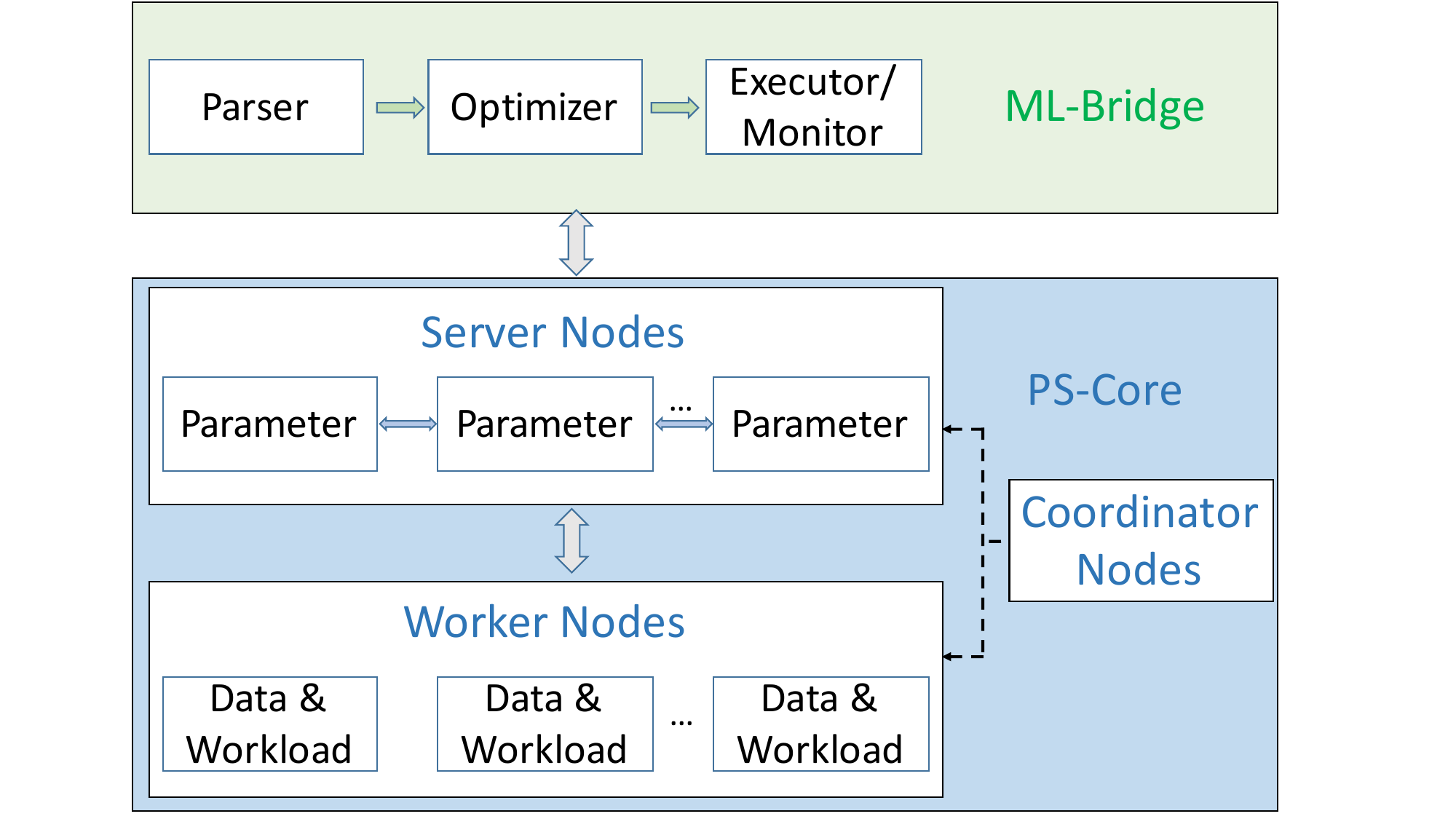}
	\caption{The simplified architecture of KunPeng, including ML-Bridge and PS-Core.}
	\label{kunpeng_arch}
\end{figure}

To make it more convenient to use, ML-Bridge, which is a practical machine learning (ML) pipeline, is provided upon the core part of KunPeng, so that the users can conveniently use the system by writing some simple scripts. The simplified architecture of the whole system is illustrated in Figure \ref{kunpeng_arch}, with two main parts showed: the ML-Bridge and the PS-Core. The users only need to operate on the ML-Bridge layer.

\subsection{Multiple Additive Regression Tree}
In this section, we will briefly introduce Multiple Additive Regression Tree (MART), and address some important features of MART which are utilized in our implementation of the distributed deep forest framework. Then, the distributed implementation of MART in KunPeng is briefly explained. Since MART is already implemented in KunPeng with great efficiency and effectiveness, and significant improvements have been shown to be reached by making it as the basic learners, we will use it as the basic building block for the distributed deep forest implementation, and any other kinds of building blocks can also be employed for the distributed version of the deep forest.

Multiple Additive Regression Tree (MART), which is also known as Gradient Boosting Decision Tree (GBDT) and Gradient Boosting Machine (GBM)~\cite{friedman2001greedy}, is a widely used machine learning algorithm in both academic and industrial fields, because of their high effectiveness and great interpretability~\cite{zhou2017kunpeng}. In fact, most of the winning Kaggle competitions or data science projects always use an ensemble of MART or its variants~\cite{chen2016xgboost} as the final model, due to its superior performance. 

To give a quick glance of MART,  we will first briefly introduce boosting decision tree \cite{drucker1996boosting}. Boosting decision tree constructs the model by additively fitting a tree model to the current residual. Let $\bm{x}_i$ denotes the $i$-th instance and $y_i$ is the corresponding label of this instance. At the $t$-th iteration, we have the prediction $F_{t-1}$ from the first $(t-1)$-rounds, then the tree model $f_t$ is learned to minimize the following objective,
\begin{equation}
\label{objective-1}
L^t=\sum_{i=1}^{n} l(y_i,F_{t-1}(\bm{x}_i)+f_t(\bm{x}_i)) + \Omega(f_t) \,,
\end{equation}
in which $l$ is the loss function and $\Omega$ is the regularization term which controls the complexity of the tree model $f_t$. Then, the prediction $F_{t}$ from the $t$-round is
\begin{equation}
\label{prediction-t}
F_{t}(\bm{x}_i) = F_{t-1}(\bm{x}_i) + f_t(\bm{x}_i)\,.
\end{equation}

Similar to boosting decision tree, MART is also additively constructed with the philosophy of fitting the residual. However, in boosting decision tree, finding the best model $f_t$ for an arbitrary loss function $L$ may get to be computationally infeasible in many conditions.  To handle this, MART \cite{friedman2001greedy} is proposed,  by making an approximation to the real residual with the steepest-descent method, the so-called pseudo-residual is fitted. Furthermore, second-order approximation is widely explored to efficiently optimize the objective \cite{friedman2000additive}, and it has been implemented in most of the systems such as XGBoost \cite{chen2016xgboost}  and lightGBM \cite{ke2017lightgbm}. The objective can be shown as below,
\begin{equation}
\label{objective-2}
\hat{L}^t=\sum_{i=1}^{n} [l(y_i,F_{t-1}(\bm{x}_i))+g_if_t(\bm{x}_i)+\frac{1}{2}h_if^2_t(\bm{x}_i) ]+\Omega(f_t) \,,
\end{equation}
in which $g_i=\partial_{F_{t-1}}l(y_i,F_{t-1}(\bm{x}_i))$ and $h_i=\partial^2_{F_{t-1}}l(y_i,F_{t-1}(\bm{x}_i))$ are the first order and second order gradient on the loss function.

Many scalable systems, such as XGBoost~\cite{chen2016xgboost} and lightGBM~\cite{ke2017lightgbm}, are some well-known implementations for this model and its variants, with additional optimization of the speed and memory usage. Similarly, KunPeng-MART is the parameter server based implementation on KunPeng system. Currently, many machine learning tasks in Ant Financial and Alibaba is using KunPeng-MART in a daily basis including many predictive tasks involving machine learning models during the double 11 online shopping festival and other daily online financial services~\cite{double11}.

There are two particular features for MART that are worth noticing and important for further using in our system. We will give a briefly discuss below.

\textbf{First}, severe class-imbalance data is often encountered in various tasks including fraud detection~\cite{chan1999}, anomaly detection~\cite{chandola2009anomaly}, and medical diagnosis~\cite{kononenko2001machine}, etc. Take two-class classification as an example, the number of some class of data may be seriously less than that of the other class of data. If we simply use the common strategy without special design, the performance may be pretty unsatisfactory~\cite{japkowicz2002class}. To handle the class-imbalance problem, two different solutions are always employed, i.e., the cost-based methods~\cite{zhou2006training} and the sampling-based methods~\cite{liu2009exploratory}. 

Note that the cost-based strategy \cite{zhou2006training} can be naturally embedded to MART, and thus in our implementation, we use cost-based strategy to deal with the class-imbalance problem. Concretely, the weight is assigned to each sample, and higher weights will be set to the class with less amount (which are always the ones with larger cost if they are misclassified) while smaller weights are set to the class with more amount (which are always the ones with smaller cost if they are misclassified). Thus, the goal in Equation~\ref{objective-2} can be modified as following,
\begin{equation}
\label{objective-3}
\hat{L}^t_w=\sum_{i=1}^{n} w_i[l(y_i,F_{t-1}(\bm{x}_i))+g_if_t(\bm{x}_i)+\frac{1}{2}h_if^2_t(\bm{x}_i) ]+\Omega(f_t) \,,
\end{equation}
in which $w_i$ is the importance weight associated with instance $\bm{x}_i$.

\textbf{Second}, extremely high dimensional data is ubiquitous in industrial tasks, and feature selection \cite{dash1997feature} is always a necessary and important step in the whole pipeline. Fortunately, the estimates of feature importance can be calculated by MART and feature selection can be performed \cite{xu2014gradient}. Generally speaking, the importance score of each attribute indicates the importance of it when constructing a tree. The more frequently a feature is used to make a decision, the more important it is. Concretely,  for every single tree,  the importance of an attribute $j$ is calculated by,
\begin{equation}
\label{importance-each}
\hat{I}^2_j(T) =  \sum_{l=1}^{L-1} \hat{i}_l^2 \mathbb{1} (v_l=j)
\,,
\end{equation}
in which $L$ is the number of leaf nodes (and $L-1$ is the number of non-terminal nodes), $v_l$ is the corresponding feature associated with node $l$, and $\hat{i}_l^2$ is the corresponding empirical improvement in square-error from the splitting and $\mathbb{1}$ is the indicator function.
Then, as shown in equation\ref{importance-global}, the global importance $\hat{I}^2_j$ of an attribute $j$ is calculated by averaging the importance value $\hat{I}^2_j(T_m)$ that the feature obtained from each single tree \cite{friedman2001greedy}.
\begin{equation}
\label{importance-global}
\hat{I}^2_j = \frac{1}{M} \sum_{m=1}^{M} \hat{I}^2_j(T_m)
\,.
\end{equation}

To handle the industrial tasks, the distributed version of the MART model is implemented in our distributed system KunPeng, which is named KunPeng-MART. Currently, many machine learning tasks in Ant Financial and Alibaba is using KunPeng-MART on a daily basis. 
There are many challenges encountered when developing distributed MART, such as the storage problem and computation and communication cost \cite{zhou2017kunpeng}. To meet the need for extremely huge storage, a data parallelization mechanism is employed in KunPeng-MART. To be specific, each worker only stores a subset of the whole data for each feature, and the main workflow for splitting a node is as follows: (1) Each worker calculates the local weighted quantile sketch with the data stored on it; (2) Each worker pushes the local weighted quantile sketch to servers, and the servers merge them up to a globally weighted quantile sketch, and find the splitting value; (3) Each worker pulls the splitting value from servers and splits samples to two nodes. Another key challenge is that the computation and communication cost of the split-finding algorithm may become very high. To handle this, the communication schema of KunPeng is employed to reduce the cost of merging local sketches, and this really speeds up the whole process.
We will use Kunpeng-MART as the building block for the distributed version of the deep forest, and employ the aforementioned features of MART when handling our tasks.

\begin{figure}[htpb]
	\centering
	\includegraphics[width=12cm,trim={27cm 5cm 0cm 0cm},clip]{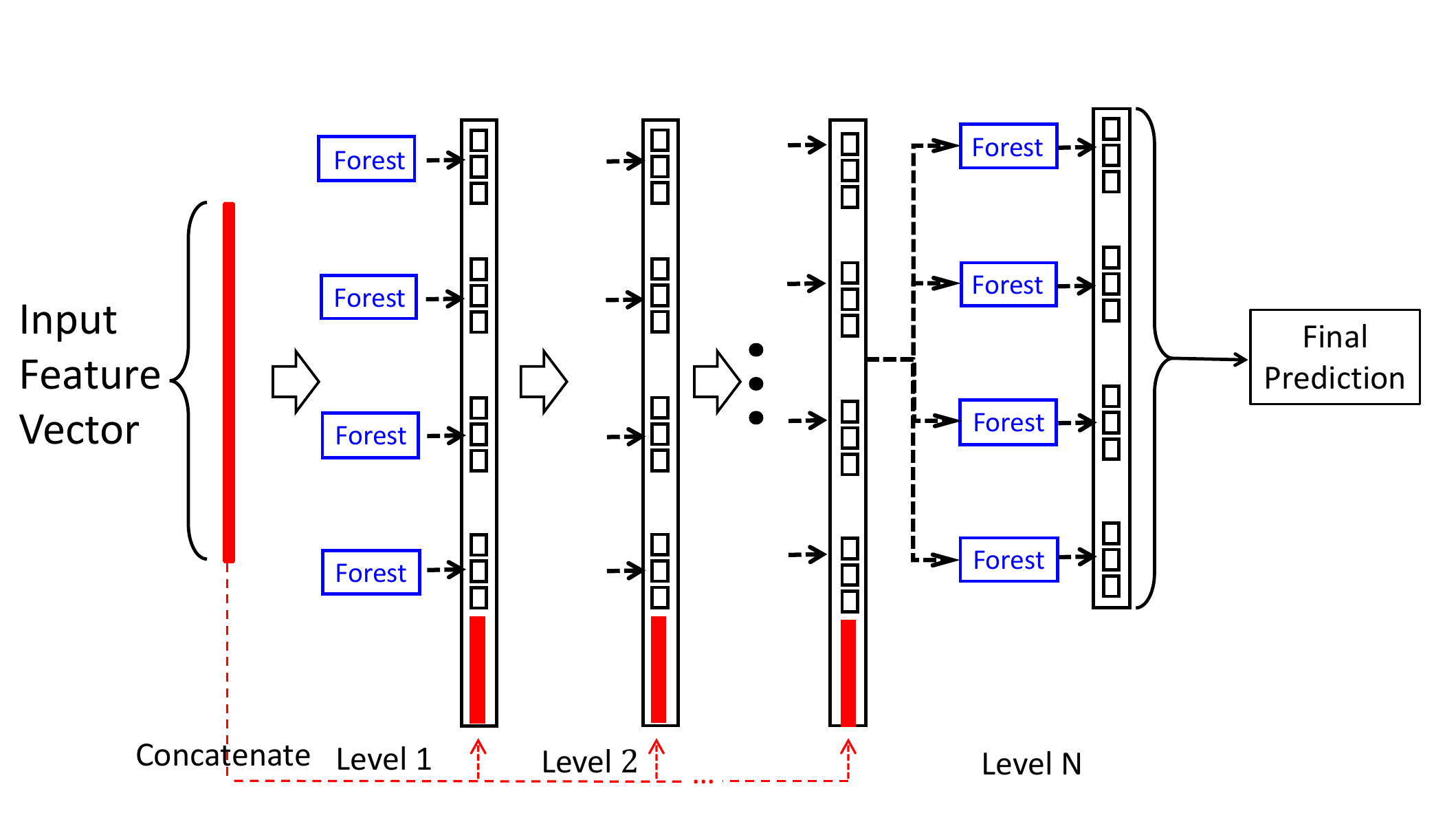}
	\caption{The cascade module of the original deep forest model~\cite{zhou2017deep}. Suppose each layer consists of four forests and there are three classes to predict; thus, each forest will output a three-dimensional vector, and each layer will output a twelve-dimensional vector, which is then concatenated with the original feature vector as the new representation, and input to the forest in the next layer.}
	\label{deepforest_cascade}
\end{figure}

\subsection{The Specified Deep Forest Structure}
In this section, start by a brief introduction of the original deep forest model, our specified version of deep forest model is introduced. We bring many improvements to the original deep forest model for the consideration of better performance for real industrial tasks, which will be explained in detail below.

Deep forest \cite{zhou2017deep,pang2018improving,zhou2018deep} is a recently proposed deep learning approach which uses tree ensembles ~\cite{zhou2012ensemble} as its building blocks in each layer. 
The original version consists of two modules, i.e., the fine-grained module and the cascading module. As discussed in the original paper~\cite{zhou2017deep}, when there are spatial or sequential feature relationships, the multi-grained scanning process helps improve performance apparently. In our task, the data is not spacial or sequential, so the fine-grained module is removed to achieve better efficiency.

We focus on the cascading module for our implemented system.
As shown in Figure \ref{deepforest_cascade}, in the cascading module of the original deep forest model, a multi-layer structure is built. Each layer can be regarded as an ensemble of ensemble, which consists of several base learners, and each base learner is an ensemble of decision tree forests, i.e., a random forest \cite{breiman2001random} or a completely random tree forest \cite{liu2008isolation}. Here, different types of forests are used for the purpose of improving diversity~\cite{sun2018structural} so that the overfitting problem can be alleviated. 

The model is built layer by layer. Each layer receives the processed feature information outputted from its preceding layer, as well as the original input features of the data (the red part in the Figure \ref{deepforest_cascade}), and concatenate them together as the input features for the forests in this layer (the first layer uses only the original feature), then the processing results of all these forests in this layer are  outputted to the next level. The processing result of one level is the combination of the class vectors generated by its forests. For example, if $m$ base learners are trained in each layer for a $k$-classes task, each forest will output a $k$-dimensional vector, and the output of each layer is a ($m*k$)-dimensional vector for each sample, and this vector will be concatenated with the original feature of this sample as the input feature for the base learners in the next layer. The method of passing the output of one layer's base learners as input to the next layer is related to stacking ~\cite{breiman1996stacked,wolpert1992stacked}.
As suggested in ~\cite{zhou2012ensemble,ting1999issues}, K-fold cross-validation is conducted in each layer to reduce the risk of overfitting. After expanding a new layer, the performance of the whole cascade can be estimated on a validation set, and the cascading process will terminate when the accuracy on validation set stops increasing; thus, the number of cascade layers is automatically determined in the deep forest.

\begin{figure}[htpb]
	\centering
	\includegraphics[width=14cm,trim={12.5cm 5cm 0cm 0cm},clip]{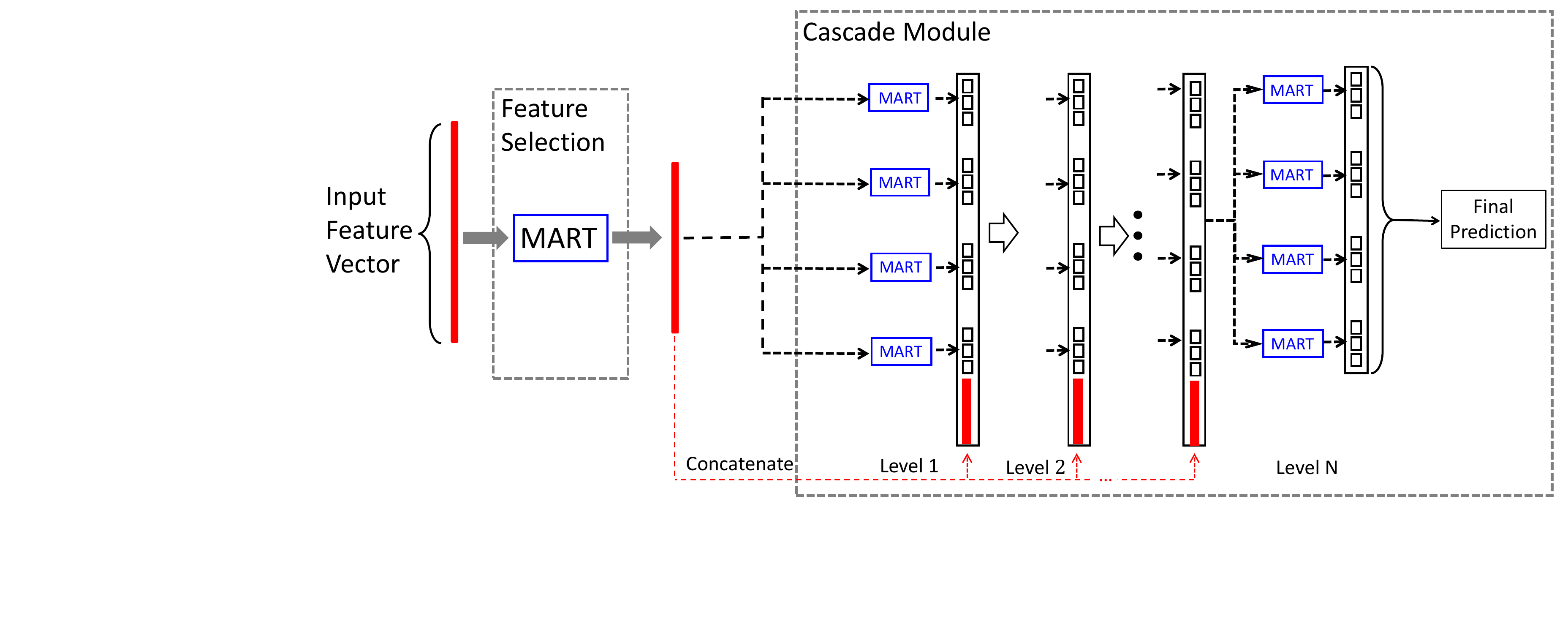}
	\caption{The pipeline of the specified deep forest structure, which includes feature selection module and cascade module. MARTs are used as the base learners.}
	\label{deepforest_overall}
\end{figure}

To meet the need for the applications in Ant Financial, the original version of deep forest may not be enough, and we develop our specific version of the deep forest pipeline, which is shown in Figure \ref{deepforest_overall}.
There are some challenges that should be taken into consideration when building a statistical model for real-world tasks.

\textbf{Firstly}, the raw training data is often in high-dimensional space, usually, thousands or even more raw features are used to represent a single entity or a transaction, in which many irrelevant attributes may exist. This can make the training process really time-consuming and resource -consuming. In addition, when deploying a model for real-time prediction, it is not economically efficient and unnecessary to calculate every attribute for each prediction. To handle this problem, the raw training data is first trained on a MART for feature importance evaluation. Then, based on the feature importance score, feature selection is conducted for the training set. The feature dimension can be greatly reduced through this procedure, while the performance will not dramatically degrade so that the filtered features can be used for further model training.

\textbf{Secondly}, the data we are facing is often extremely class imbalanced, the positive samples may be much rare, compared with negative samples. Sometimes, the number of negative samples can be as much as 10 thousand times of the number of positive ones. Therefore, a mechanism which can effectively handle such a situation is needed in order to get a reasonable result. To deal with this problem, the cost-based strategy is employed in each base learner so that the class-imbalance problem is addressed. Concretely, samples from different classes will be allocated with different weights, while the positive ones (which are also the minority) are always with bigger weights, meaning that the misclassification of them will lead to a larger loss.

\textbf{Thirdly}, for the extremely large scale tasks in industrial settings, both the data size and the features size can be huge, and the running time may be pretty lengthy, so efficiency and effectiveness are both important. To meet this, all base learners in the original deep forest (i.e., random forests and completely random tree forests) are replaced with the MART model implemented in our Kunpeng system, which is shown to be able to provide excellent performance with consideration of both efficiency and effectiveness. What's more, the supplement of ~\cite{zhou2017deep} also shows that a significant improvement can be reached by using MART as the base learners.

\textbf{Finally}, for many tasks in the industrial scenario, the evaluation metric may be specific, so the original applied metric accuracy in the deep forest cannot meet all of the need (for example, accuracy may not be a good metric for the severely imbalanced data). To handle this problem, we provide more metrics (such as AUC \cite{fawcett2006introduction} and F1-Score \cite{powers2011evaluation})  for the automatic growing of the cascading structure. What's more, the specifically designed evaluation metrics can also be introduced for personalized demand.

Furthermore, note that we use MART as the base learner of deep forest since it is already implemented in KunPeng, and better efficiency and effectiveness can be obtained in this way, while in the original paper, random forest \cite{breiman2001random} and completely random tree forest \cite{geurts2006extremely,liu2008isolation} is used as the base learner to provide better diversity (which is crucial for ensemble methods) \cite{zhou2012ensemble,tang2006analysis,kuncheva2003measures},  when replacing all base learners with MART, the diversity is damaged to some extent, so some strategies are applied in MART to alleviate this problem, including instance sampling, setting different number of trees for each forest and setting different depth for trees in different forests.

It should be noticed from the above description, each base learner can be trained in a distributed fashion, and all the base learners within a layer can also be trained in parallel, making the whole process easy to be implemented in a distributed fashion. The next section will briefly explain how to build such a model via a parameter server based distributed learning system.

\subsection{Distributed Implementation and Job Scheduling}
In industrial scenarios, we are always in confront of the data with tremendous size and high dimension, which means that the distributed version of the algorithm is needed. On the other hand, the deep forest model has been proved to be effective in a range of different tasks; however, the performance of this model has not been verified in extreme-large scale tasks. In this section, we introduce the distributed framework and job scheduling method for the development of the distributed version of the deep forest model.

The distributed deep forest framework is built upon the widely used parameter server based system KunPeng in Alibaba and Ant Financial. Based on the KunPeng architecture, the distributed version of deep forest contains three different kinds of nodes: (1) the worker nodes, which execute the heavy computing tasks; (2) the server nodes, which maintain the  globally shared parameters; (3) the coordinator nodes, which coordinate the worker and server nodes, and perform job scheduling for the whole task.

To deploy the deep forest algorithm in a distributed manner, a key problem is how to do job scheduling for the whole pipeline.
In each layer, the process of deep forest consists of the following sub-jobs: (1) data preparation process, which splits the whole dataset into different training and valid fold, since k-fold cross validation is needed to reduce the risk of over-fitting; (2) model training process, which trains different base learners based on the split training data parallelly; (3) prediction process, which makes prediction on the split valid data parallelly; (4) combination and concatenation process, which combines the output of different base learners from the previous level and concatenates these output with the original features, and this processed data will be used as the input feature for the subsequent level.

\begin{figure}[htbp]
	\centering
	\includegraphics[width=11cm,trim={4cm 0cm 4cm 0cm},clip]{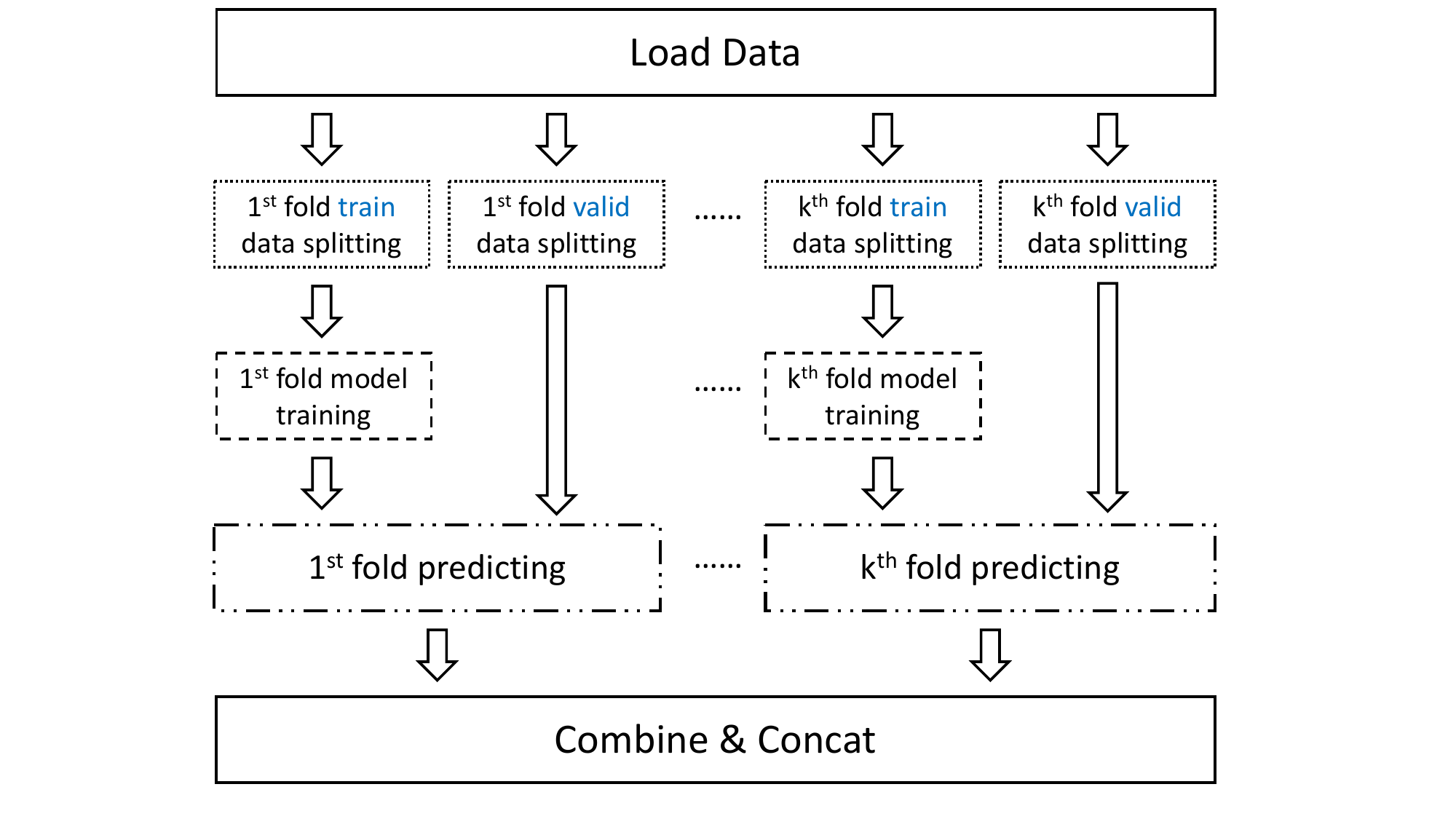}
	\caption{Job scheduling with DAG, each rectangle represents a process, only the corresponding processes are connected with each other.}
	\label{dag}
\end{figure}

To perform efficient job scheduling, the directed acyclic graph (DAG) \cite{thulasiraman2011graphs} is employed in our system. A directed acyclic graph is a finite directed graph with no directed cycles. As shown in Figure \ref{dag},  we regard each aforementioned process as a node in the graph, and only the corresponding process are connected. The pre-conditions of one node are the input of that node. Only when all the pre-conditions of one node are satisfied, that node could be executed. 
What's more, each node is executed separately, which means that the failure of one node will not influence other nodes.
Each node is dependent on its inputs, which means that if and only if all the pre-conditions of the node are finished, that node is allowed to execute.
In this way, the node will not wait for the performing of irrelevant nodes, and the waiting time will be significantly shortened since each node only needs to wait for the corresponding per-nodes finishing executing. For example, once the splitting of the $k$-fold training data is finished, the corresponding model on this fold can be called to do training, rather than waiting until all of the data are prepared, and the prediction on the related valid data will be called to execute if the training is finished. Besides, this design provides a better solution for failover. For example, if some node is crashed for some reason,  we will only need to rerun from this node instead of running the whole algorithm from the beginning since its pre-conditions are finished successfully.

\subsection{Graphical User Interface}
Data scientists in Ant Financial and Alibaba are facing hundreds of different machine learning tasks each day. Numerous new tasks are created and each task is different by its own nature. Therefore, how to efficiently build and evaluate a model is critical for productivity. In order to solve such problem, in Ant Financial, the Platform of Artificial Intelligence (PAI for short) \footnote{pai.alipay.com}  is developed, which decouples the algorithms from different algorithm engines, for example KunPeng, MaxCompute, MPI, etc., and provides a uniform graphical user interface (GUI) for data scientists to process data, invoke multiple machine learning algorithms, create task pipeline at cloud, and so on.

\begin{figure}[htbp]
	\centering
	\includegraphics[width=10cm,trim={0cm 0cm 0cm 0cm},clip]{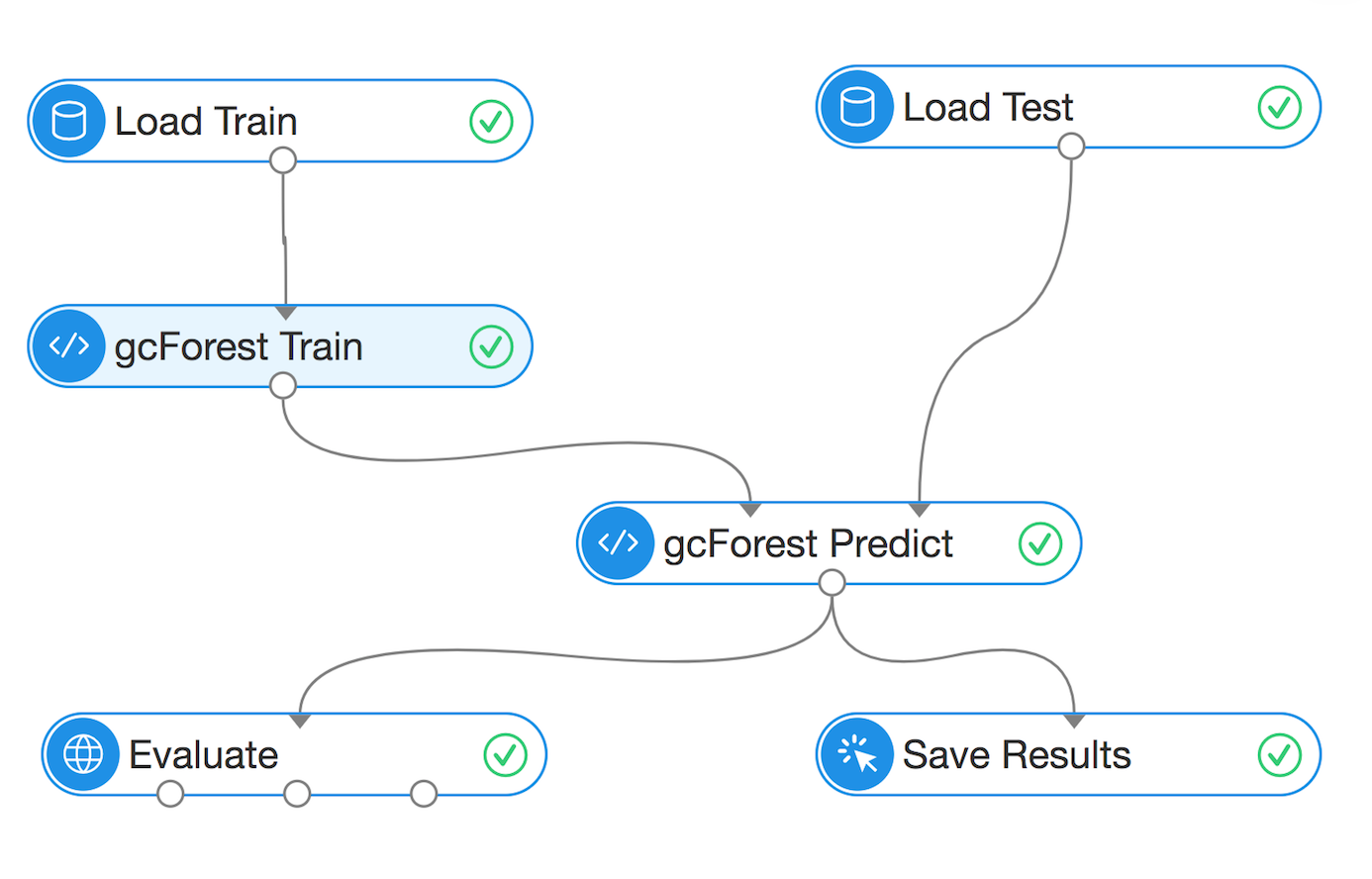}
	\caption{The overall GUI of deep forest on PAI, each node represents an atomic operation.}
	\label{GUI}
\end{figure}

As noted earlier, the deep forest algorithm is robust enough to handle different domains of tasks, making it one of the best choices when facing a new task. The parameter server based implementation of base leaner enables the model to handle even extremely large scale real-world problem, and we have implemented a deep forest module in PAI platform, the data scientists are able to create a deep forest model within a browser. That is, with only a few clicking of the mouse, the deep forest model is ready to train on massive training data and ready for deployment.

\begin{figure}[htbp]
	\centering
	\includegraphics[width=13cm,trim={0.1cm 13cm 0cm 0cm},clip]{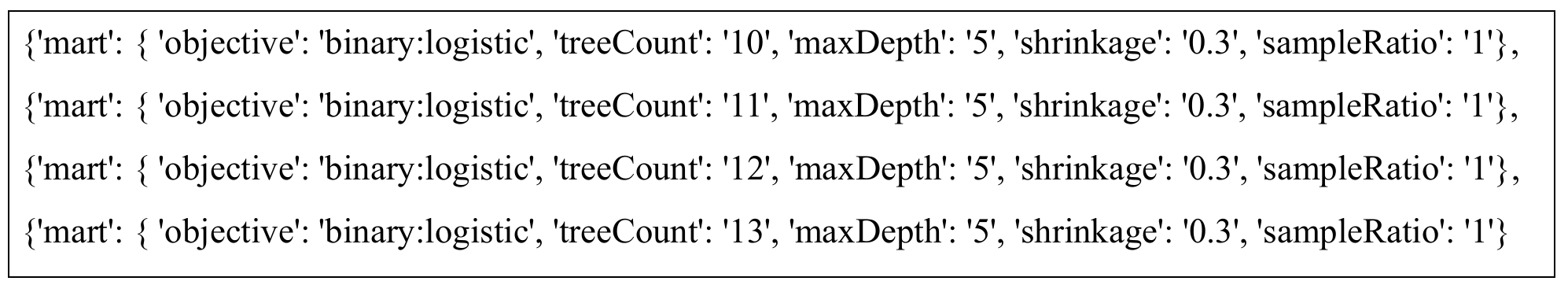}
	\caption{A simple example of the configuration for the training process.}
	\label{config}
\end{figure}

The demo of an overall GUI of the process with a deep forest model is illustrated in Figure~\ref{GUI}. Each node represents an atomic operation, which includes loading the data, building the model, making predictions and performing the evaluation, etc. All the details of the deep forest model are encapsulated into a single node, the only thing needed to specify is which base learner to use, how many base learners per layer and the detailed configuration of each base learner, as shown in Figure~\ref{config}. The default base learners are MART, as introduced before.

The arrowed line indicates the sequential dependency and data flow from one to the other. By only a few drag-and-clicks, the user is able to create the deep forest model within minutes, and the evaluation results will be analyzed once the training is finished. If a deep forest is determined to be deployed, the trained model can be directly inputted to the data awaiting assessment.

\section{Application}

In this section, we validate the effectiveness of deep forest model on one important application for Ant Financial, namely the automatic detection of cash-out fraud. We will first give a detailed introduction to the task of cash-out fraud and a brief introduction of the data preparation process, then the empirical results are showed and analyzed from different perspectives. What's more, the robustness of the deep forest model is validated.

\subsection{Task Description}

Cash-out, which means pursuing cash gains using illegal or insincere means, is a troublesome problem in the scenario of the credit card, public accumulation funds and credit financial companies like Ant Financial.  
Similar to the credit card, Ant Credit Pay is a credit production provided by Alipay. Alipay provides each user with a certain amount of credit, and the user can consume with the credit but not money to buy anything using Alipay and pay back later, which is pretty similar to the use of credit card.
However, there are always some people who try to pursuing illegal gain from these credit productions, and this becomes a serious threat for companies like Ant Financial.
Thus, the automatic detection of the cash-out fraud for Ant Credit Pay and the corresponding treatment strategy for the detected potential cash-out transactions are pretty crucial for risk control in these companies. 

The cash-out fraud by Ant Credit Pay always follows the following process: the shopper makes a transaction to the seller with Alipay by scanning a QR code, and pay to the seller using Ant Credit Pay. In this way, the credit but not the money of the shopper is consumed and the shopper can get cash from the seller. The threat is that the shopper will not pay back the money later, which will result in an economic loss for the company.
And also, similar activities like the cash-out fraud may give a potential threat to the credit system. 
Without a  proper strategy to detect them, millions of CNY may lose each day. 
What we need to do is to provide the system with the ability of detecting the potential hazard of cash-out when a QR code is scanned for a transaction, and the system can automatically disable the consuming of the credit in Ant Credit Pay for this payment if the transaction is detected to be a highly possible cash-out fraud, so that the shopper can only consume with money but not the credit; thus, the economic loss can be avoided.

\subsection{Data Preparation}

We formulate this task as a binary classification problem to predict if a transaction is a potential fraud transaction, and collect the original features from four different aspects, i.e., the seller features, which describe the identity information of the seller (such as the transaction amount information); the buyer features, which describe the identity information of the buyer (such as the gender and age information); the transaction features, which describe the information of the exact transaction (such as the time of the transaction), and the history features, which describe the information of both seller and buyer (such as the history transaction amount and other statistics in a period). 
Altogether, more than 5000 dimension features are collected when each transaction is happening, with both numerical and categorical features included, so this problem is a hybrid modeling problem with huge feature size.
What's more, we need to address that this task is naturally class-imbalanced, since the amount of fraudulent transactions is only a tiny percentage of the amount of all transactions.

To construct the training and test data, we sampled the training data from data collected from O2O (\textbf{O}nline to \textbf{O}ffline)  transactions using Ant Credit Pay during some successive months, and the sampled transactions of the same scenario during the next few months are used as test data.
The statistical information of the data is shown in Table \ref{data_info}. As we can see, this task is with extremely large scale (with more than 131 millions of samples for training) and is severely class-imbalanced (the number of negative instances is more than 700 times of the number of positive instances in training set).

\begin{table}[htbp]
	\centering
	\caption{The number of the training and test instances (including the number of positive instances, negative instances and all instances).}
	\label{data_info}
	\scalebox{1.0}{
		\begin{tabular}{|p{1.2cm}<{\centering}|p{1.5cm}<{\centering}|p{1.5cm}<{\centering}|p{1.5cm}<{\centering}|}
			\hline
			& \# Pos. Ins. & \# Neg. Ins.  & \# All Ins. \\
			\hline
			Train  &  171,784 & 131,235,963  & 131,407,704   \\ 
			Test  &  66,221  & 52,423,308  & 52,489,529  \\
			\hline
		\end{tabular}
	}
\end{table}

As we discussed above, the original collected features are with the size up to 5000, among which many irrelevant attributes may be included. If simply use all these raw features, the whole procedure may get to be too time-consuming and resource-consuming, and it will really reduce the efficiency when deployed as a real-time service, making it a bad choice. On the other hand, according to the business experience, the size of the finally-used features can be greatly reduced without significant degradation of the performance. To bypass this obstructor, As shown in Fig.~\ref{deepforest_overall}, before training the final models, feature importance is firstly calculated and feature selection is performed with the help of MART. 

Concretely, We first run a MART model with all collected features and then feature importance scores are calculated with the obtained model using Equation~\ref{importance-global}. 
Based on the obtained feature importance scores, feature selection is then performed. We evaluate the performance with different selected feature size. Empirical results show that with the first 300 selected features (which are with the higher feature importance scores), the retrained model can already achieve the competitive performance with the model trained using the whole features, which also validates the redundancy of the raw features. Thus, the subsequently-used features are filtered to be the 300 dimensions with higher importance scores and the whole efficiency is greatly improved.

\subsection{Empirical Results}
In this section, empirical results are shown from different perspectives and a detailed discussion is provided below.

\subsubsection{Experiment Setting}
Since the data is on an extremely large scale, the experiments are conducted with the distributed learning system KunPeng.
To validate the effectiveness of the deep forest algorithm on this extremely large dataset, many KunPeng based algorithms are compared, including logistic regression (LR) \cite{liu2009large},  deep neural network (DNN) \cite{goodfellow2016deep} and multiple additive regression trees (MART) \cite{friedman2001greedy}, in which LR and MART are the previously and currently deployed machine learning based methods, which have achieved some success during the past period. Since the problem is a hybrid modeling problem, we believe that DNN may not be a good choice for this task, and experiments are conducted with DNN to validate the effectiveness. Note that since the data is severely imbalanced, the cost-based strategy is employed in these models, i.e., instances from the positive and negative class are allocated with different weights. We need to address that, for MART, 600 trees are used to get better performance (and a larger number of trees will not further improve the performance), while in the deep forest, each MART is with at most 200 trees, but not the same number of 600.
For DNN, following the procedure explained in ~\cite{zhou2017deep}, we examine a variety of architectures with a validation set, and pick the one with the best performance, and retrain the whole network to get the final result. Specifically, the activation function is selected from ReLU and sigmoid, the unit number of each hidden layer is selected from {100,300,500,1000}, and the dropout rate is selected from 0.25 and 0.5.

\subsubsection{Common Metrics}: We first evaluate the performance with the widely used metrics for binary classification tasks in Ant Financial, including the AUC (\textbf{A}rea \textbf{U}nder the ROC \textbf{C}urve), the F1 score and the KS (\textbf{K}olmogorov-\textbf{S}mirnov) score . 
These metrics will not directly reflect the economic influence by each model, but even a slight improvement at the third decimal place for these metrics is significant since this improvement can bring into millions of decrease of economic loss, according to the business experience, so we first give the result and analysis from these metrics.

The results are shown in Table \ref{result_1}, as we can see, the deep forest method (gcForest for short) performs much better than all other existing methods, the advantage is especially noticeable for F1-Score since the AUC and KS-Score are already pretty high for other methods. 
MART performs as the second better, validated its effectiveness. We need to address that the MART model is fine-tuned, and 600 trees are needed to reach this performance (and a larger number of trees will not further improve the performance), which may be the upper limit that can be reached by this model.
However, for deep forest, only with 200 trees for each MART, and with slight tuning, the performance is already much better than the best baseline, and as we will show later, even with only 50 trees (the default setting) for each MART in deep forest, the performance is still better than the fine-tuned MART model. 
Here, DNN performs pretty unsatisfactorily, which verified the weakness of DNN for handling the hybrid modeling problem (while DNN is more suitable for the continuous modeling tasks like image recognition). 
LR performs unsatisfactorily too when comparing to MART and deep forest.

\begin{table}[htbp]
	\centering
	\caption{The results using the common metrics, i.e., AUC, F1-Socre and KS-Score.}
	\label{result_1}
	\scalebox{1.0}{
		\begin{tabular}{|p{1.8cm}<{\centering}|p{1.5cm}<{\centering}|p{1.5cm}<{\centering}|p{1.5cm}<{\centering}|}
			\hline
			& AUC  & F1-Score   & KS-Score  \\
			\hline
			LR  &  0.9887   & 0.4334   & 0.8956    \\ 
			DNN  &  0.9722   & 0.3861   & 0.8551   \\
			MART & 0.9957 &   0.5201   &  0.9424      \\ 
			gcForest & \textbf{0.9970}  & \textbf{0.5440}   & \textbf{0.9480}     \\
			\hline
		\end{tabular}
	}
\end{table}

\subsubsection{Specified Metrics}: Besides the common metrics, as a real-world task, many specified metrics are always needed for analysis, with the consideration of practical using. For our task, one important metric is the recall of the positive samples (the potential cash-out cases) under a given interrupt (by disabling Ant Credit Pay for such transactions) rate. For example, if we can stand that 1/10000 of all transactions to be interrupted, we will definitely hope that the selected cases contain as many positive samples as possible. 
When deploying these models, a threshold is always determined so that some ratio of the transaction will be interrupted. Since the volume of transaction is tremendous, a small improvement under this metric will be associated with a large number of transactions and a large amount of money can be saved for the company.

In Table \ref{result_2}, the recalls under different interrupt rates (for example, 1/100 means that 1/100 of all transactions are interrupted) are provided. As we can see from the results, LR and DNN perform pretty unsatisfactorily, the MART brings great improvement from these methods.
At any level of interrupt rate, the deep forest model can always capture most potential cash-out transactions, which means that it will provide the best performance when deployed. The improvements can be bigger than 2\% even compared with the fine-tuned MART under the interrupt rate of 1/10000 and 1/1000, which is the currently deployed model for this task.  The improvement for these metrics are much significant according to the business specialist, and these metrics are much more relevant to the real economic influence when the model is deployed.
\begin{table}[htbp]
	\centering
	\caption{The results using the specified metrics, i.e.,the recall of the positive samples (the potential cash-out cases) under a given interrupt rate. Here, 1/100 means that 1/100 of all transactions are interrupted.}
	\label{result_2}
	\scalebox{1.0}{
		\begin{tabular}{|p{1.8cm}<{\centering}|p{1.5cm}<{\centering}|p{1.5cm}<{\centering}|p{1.5cm}<{\centering}|}
			\hline
			& 1/10000 & 1/1000  & 1/100 \\
			\hline
			LR  &  0.3708   & 0.5603   & 0.8762    \\ 
			DNN  &  0.3165   & 0.4991   & 0.8471   \\
			MART & 0.4661 &   0.6716   &  0.9358      \\ 
			gcForest & \textbf{0.4880}  & \textbf{0.6950}   & \textbf{0.9470}     \\
			\hline
		\end{tabular}
	}
\end{table}

\subsubsection{PR curve}: Furthermore, according to the business experience, PR (\textbf{P}recision-\textbf{R}ecall) curve is always a good choice to provide a visual comparison, and the decision is always made based on this metric, so the improvement in this metric is more approved. A PR curve shows the relationship between precision and recall for each possible cut-off. Generally speaking, the curve with the bigger covering area is always be regarded as a better one when comparing to the one with a smaller coverage area.

\begin{figure}[htbp]
	\centering
	\includegraphics[width=11cm,trim={0cm 0cm 0cm 0cm},clip]{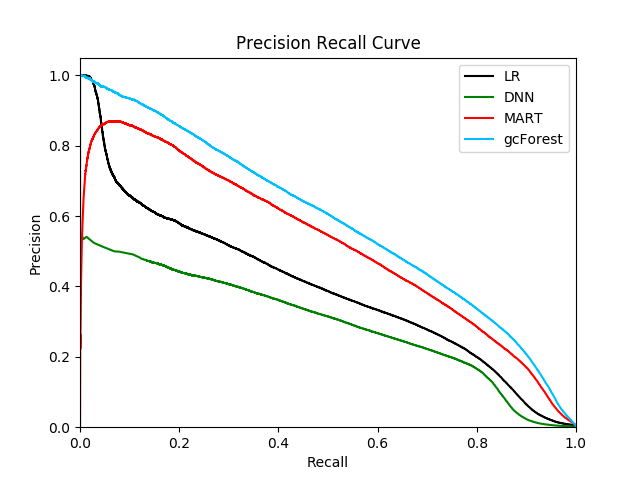}
	\caption{The PR curve of LR, DNN, MART and gcForest.}
	\label{pr_curve_1}
\end{figure}

We draw the PR curve of different tested methods in Figure \ref{pr_curve_1}, the result is much clearer, the PR curve of deep forest covers those of all other methods, meaning that the performance of deep forest is much better than other methods, validating the effectiveness of deep forest model. One interesting result is that the MART turns out to behave unsatisfactorily in the high score part, making it somewhat unsatisfactory for deployment.

\subsubsection{Economic benefit}: Till now, we have provided analysis from different perspectives, and all of these results validate the effectiveness of the deep forest model on this extremely large scale task. Furthermore, when deployed, the deep forest model can block cash-out fraud transactions in a large amount of money every day (the detail is business confidential). Even compared with the best-deployed model (MART with 600 trees), the deep forest model (200 trees for each MART) can still additionally bring into a significant decrease of economic loss each month, making it a better choice for this task.

\subsection{Robustness Analysis}
In this section, we give a brief discussion on the robustness of the deep forest model.

According to the original paper of deep forest \cite{zhou2017deep}, a default setting of the deep forest model will produce highly competitive results in a range of tasks, which means that less effort on parameter tuning is needed when we use this model. 
To validate the robustness of parameter setting for deep forest model in large-scale setting, we compare performances among the deep forest model with default setting (using 4 MARTs as base learner, each with only 50 trees),  the slightly-tuned deep forest model (by only changing the number of trees to 200 in each MART) and the best tuned model of the MART (with 600 trees).

\begin{table}[htbp]
	\centering
	\caption{The results using the common metrics, i.e., AUC, F1-Socre and KS-Score.}
	\label{result_3}
	\scalebox{1.0}{
		\begin{tabular}{|p{1.8cm}<{\centering}|p{1.5cm}<{\centering}|p{1.5cm}<{\centering}|p{1.5cm}<{\centering}|}
			\hline
			& AUC  & F1-Score   & KS-Score  \\
			\hline
			MART & 0.9957 &   0.5201   &  0.9424      \\ 
			gcForest-d & 0.9962 &   0.5247   &  0.9444     \\
			gcForest-t & \textbf{0.9970}  & \textbf{0.5440}   & \textbf{0.9480}     \\
			\hline
		\end{tabular}
	}
\end{table}

\begin{table}[htbp]
	\centering
	\caption{The results using the specified metrics, i.e.,the recall of the positive samples (the potential cash-out cases) under a given interrupt rate. Here, 1/100 means that 1/100 of all transactions are interrupted.}
	\label{result_4}
	\scalebox{1.0}{
		\begin{tabular}{|p{1.8cm}<{\centering}|p{1.5cm}<{\centering}|p{1.5cm}<{\centering}|p{1.5cm}<{\centering}|}
			\hline
			& 1/10000 & 1/1000  & 1/100 \\
			\hline
			
			MART & 0.4661 &   0.6716   &  0.9358      \\ 
			gcForest-d & 0.4703 &   0.6775  &  0.9397      \\ 
			gcForest-t & \textbf{0.4880}  & \textbf{0.6950}   & \textbf{0.9470}     \\
			\hline
		\end{tabular}
	}
\end{table}

\begin{figure}[htbp]
	\centering
	\includegraphics[width=11cm,trim={0cm 0cm 0cm 0cm},clip]{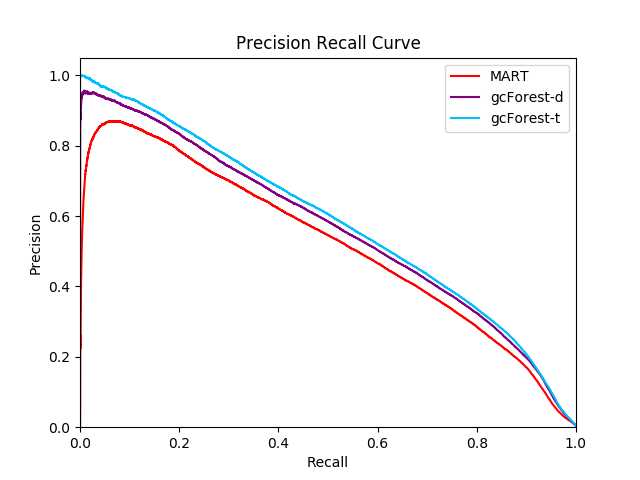}
	\caption{The PR curve of deep forest with default parameters (gcForest-d), slightly-tuned deep forest (gcForest-t) and  fine-tuned MART (MART).}
	\label{pr_curve_2}
\end{figure}

The corresponding results of the tested models are showed in Table \ref{result_3}, Table \ref{result_4} and Figure \ref{pr_curve_2} (gcForest-d for the default setting of deep forest and gcForest-t for the slightly-tuned deep forest). 
As we can see, even without any tuning, the performance of the default setting of the deep forest model gcForest-d (with only 50 trees for each MART) is still showed to be much better than the fine-tuned MART model (with up to 600 trees), which is consist to the claim from original paper that the default setting of deep forest can produce highly competitive performance. Note that even with 4 forests in each layer, the number of trees of the deep forest is still much smaller than that used in the fine-tuned MART, leading to a great resource saving.
On the other hand, a slightly-tuning of deep forest model gcForest-t (by only changing the number of trees to 200) has already lead to much better performance. We believe that a fine-tuning of the deep forest model will lead to a much more excellent result.

\section{Conclusion and Future work}

In this paper, we introduce the distributed version of the deep forest model, which is developed and deployed based on the parameter server system KunPeng. To meet the need for real-world applications, many improvements are introduced, which include, MART as the building block with consideration of both efficiency and effectiveness, the cost-based strategy for handling extra-imbalanced data, MART for feature selection, different evaluation metrics for automatically determining of the layer amount.  
Experiments on the task of automatic detection of cash-out fraud are performed and results are analyzed from different perspectives. All results show that the deep forest model can provide highly competitive performance, and a significant decrease in economic loss can be reached with this model even when compared with the best-deployed model.

There are still many directions which can be further explored in the future. Currently, most of the tasks that we are facing are supervised two-class classification problem, and the developed system along with the experiment in this paper is mainly focused on this setting. However, other problems, such as regression and multi-class classification tasks can be further tested to validate the effectiveness of deep forest model under an extremely large scale.
What's more, other widely studied settings, such as semi-supervised learning~\cite{chapelle2009semi}, multi-label learning~\cite{zhang2014review}, multi-instance learning~\cite{dietterich1997solving} and learning under label noise~\cite{liu2016classification} can also be probed.
For example, many real-world problems are in multi-label or multi-instance format, while tree based model has been successfully adapted to these problems~\cite{clare2001knowledge,prabhu2014fastxml,leistner2010miforests}, and we believe that with a proper strategy, the deep forest model may be naturally employed to these problems with better performance. What's more, the robustness under label noise is a non-negligible issue in reality. Previous study~\cite{ghosh2017robustness} has shown that many tree-based methods are robust to symmetric label noise under large sample size, and we think that the deep forest model has the potential to be robust for this problem. 
Besides, the importance reweighting framework is shown to be helpful in the presence of label noise~\cite{liu2016classification}.   
We will consider improving the robustness of deep forest under label noise setting with proper instance reweighting techniques for industrial tasks, and we will further validate the robustness of deep forest model with more experiments on proper real-world tasks.
Furthermore, it has been deemed as the holy grail challenge for the artificial intelligence community to combine machine learning and logical reasoning to work together~\cite{zhou2019abductive}, and compared to the combination of neural network and logic reasoning, combining deep forest model with logic reasoning may be more convenient to leverage domain knowledge.

\begin{acks}
	The authors want to thank reviewers for their helpful comments and suggestions. 
	This research was partially supported by the National Key R\&D Program of China (2018YFB1004300), the National Science Foundation of China (61751306), and the Collaborative Innovation Center of Novel Software Technology and Industrialization.
\end{acks}

\bibliographystyle{ACM-Reference-Format}
\bibliography{reference}

\end{document}